\documentclass{article}
\usepackage{iftex}
\ifPDFTeX
  \usepackage[utf8]{inputenc}
\fi
\usepackage[T1]{fontenc}
\usepackage{genai4health_2026}
\usepackage{hyperref}
\usepackage{url}
\usepackage{booktabs}
\usepackage{amsmath,amssymb}
\usepackage{graphicx}
\hypersetup{hidelinks}
\usepackage{xcolor}
\usepackage{subcaption}
\usepackage[section]{placeins}

\title{XRL4T1D: A Post-hoc Explainability Framework for Reinforcement Learning in Type 1 Diabetes Management}
\author{%
  Malay Phadke \\
  Sardar Patel Institute of Technology\\
  Mumbai, India \\
  \texttt{malay.phadke@spit.ac.in} \\
  \And
  Chirath Hettiarachchi \\
  Australian National University\\
  Canberra, Australia \\
  \texttt{chirath.hettiarachchi@anu.edu.au} \\
}

\begin{document}
\maketitle
\begin{abstract}
Type 1 diabetes (T1D) requires lifelong insulin therapy to maintain safe glucose levels. Reinforcement learning (RL) has shown promise for automated insulin dosing in FDA-accepted \textit{in-silico} studies using the UVA/Padova simulator, but these policies remain difficult to interpret in a safety-critical setting. Aggregate glucose-control metrics describe how well a controller performs, but not how it uses recent glucose and insulin history to determine each dose. We present XRL4T1D, a post-hoc explainability framework for evaluating pretrained RL insulin-dosing policies without retraining. Each policy maps a 60-min history of continuous glucose monitor (CGM) and administered-insulin measurements, sampled every 5 min, to the next insulin recommendation. 
Using SHapley Additive exPlanations (SHAP), we estimate how strongly each policy depends on recent CGM and insulin-history inputs. We first compare the relative importance of the two input channels and then summarize the insulin-history contribution over three 20-min windows motivated by the clinical concept of insulin-on-board.
We then evaluate whether the attributions reflect policy-output sensitivity under targeted input replacement. Across all 80 tested patient--policy--mask-size cases, replacing the inputs with the largest absolute attributions changes the normalized insulin recommendation more than replacing either the lowest-attribution or randomly selected inputs. At the temporal level, the highest-attribution insulin-history window is also the most output-sensitive under direct window replacement in 115/120 patient--policy--scenario configurations. KernelSHAP and TimeSHAP identify the same highest-attribution temporal window in 111/120 configurations.
We then apply the framework to PPO and G2P2C as case-study RL controllers across 20 virtual patients. Both policies rely predominantly on CGM history, with this dependence particularly strong for G2P2C. 
Within the smaller insulin-history contribution, however, the recent--middle--older profile is not consistently different between the two controllers and varies across cohorts, explainers, and explanation settings. In contrast, the global attribution rankings remain highly stable under the tested observation corruptions. 
These findings show that three questions should be considered separately when using post-hoc explanations to evaluate clinical policies: whether highly ranked inputs actually affect the policy output, whether the temporal ordering is reproducible, and whether finer controller comparisons remain stable across analysis choices.
\end{abstract}

\section{Introduction}

Type 1 diabetes (T1D) requires lifelong insulin therapy to maintain safe glucose levels. Automated insulin delivery systems use continuous glucose monitor (CGM) measurements to adjust insulin dosing, but current systems still require substantial user input and must operate under substantial physiological variability. Reinforcement learning (RL) offers a promising route toward more autonomous insulin dosing, and many RL controllers have been developed and evaluated in the UVA/Padova T1D simulator, an FDA-accepted physiological simulation environment that enables preclinical evaluation without risk to patients \cite{atkinson2014t1d,brown2019closedloop,breton2020children,visentin2014uvapadova,simglucose2018}.

Automated insulin controllers are usually compared using clinical glycemic outcomes such as time in range (TIR). These metrics are essential for evaluating glucose control, but they do not reveal how a controller arrives at an individual dosing decision. Two RL controllers (policies) can therefore achieve similar aggregate outcomes while relying on different parts of the same physiological history \cite{battelino2019tir}. This distinction is important for safety-critical clinical AI: beyond asking whether a controller performs well, we would also like to know which glucose and insulin measurements its recommendations depend on, whether the same conclusions hold across clinically relevant settings, and whether the resulting explanations remain stable when observations are noisy or incomplete.

The policies studied here operate on the same longitudinal state representation. Every five minutes, each policy receives 12 CGM measurements and 12 administered-insulin values sampled at five-minute intervals, representing the preceding 60 minutes, and outputs the insulin recommendation for the next five-minute interval. Thus, each decision depends on 24 historical inputs: 12 describing recent glucose and 12 describing recent insulin delivery.
We study how the importance assigned to these inputs varies across preprandial, early- and late-postprandial, and nocturnal periods, as well as across patient cohorts, meal scenarios, and simulated sensor perturbations. These contexts are used to evaluate and interpret the policy's dependence on its input history; they are not additional inputs to the policies.

Post-hoc explanation offers one way to examine this dependence. We use SHapley Additive exPlanations (SHAP) \cite{lundberg2017shap} to estimate how strongly each historical input contributes to a policy's insulin recommendation. We first compare how much importance the policy assigns to CGM history versus administered-insulin history. Within the insulin-history contribution, we then summarize the 12 temporal lags into recent (0--20 min), middle (20--40 min), and older (40--60 min) windows. This temporal summary is motivated by the clinical concept of insulin-on-board (IOB), which describes the continuing activity of previously administered insulin, but our measure reflects how the policy uses its recorded one-hour insulin history rather than estimating pharmacokinetic residual insulin.

SHAP values alone, however, are not sufficient evidence that a highlighted input meaningfully affects the policy output. Adjacent temporal measurements are correlated, and an explanation can appear plausible without accurately reflecting policy sensitivity. We therefore test SHAP importance against policy-output sensitivity in three ways: individual-input replacement, direct replacement of each 20-min insulin-history window, and comparison with TimeSHAP, a sequential Shapley-based explainer \cite{bento2021timeshap}.
We apply this framework to two publicly available pretrained controllers, PPO \cite{hettiarachchi2022reinforcement} and G2P2C \cite{hettiarachchi2024g2p2c}, across 20 virtual patients. These controllers serve as a case study for evaluating not only whether the explanation reflects policy-output sensitivity, but also whether the resulting conclusions remain stable across clinically relevant settings and reasonable explanation choices.

\begin{figure*}[t]

    \centering

    \includegraphics[width=0.95\textwidth]{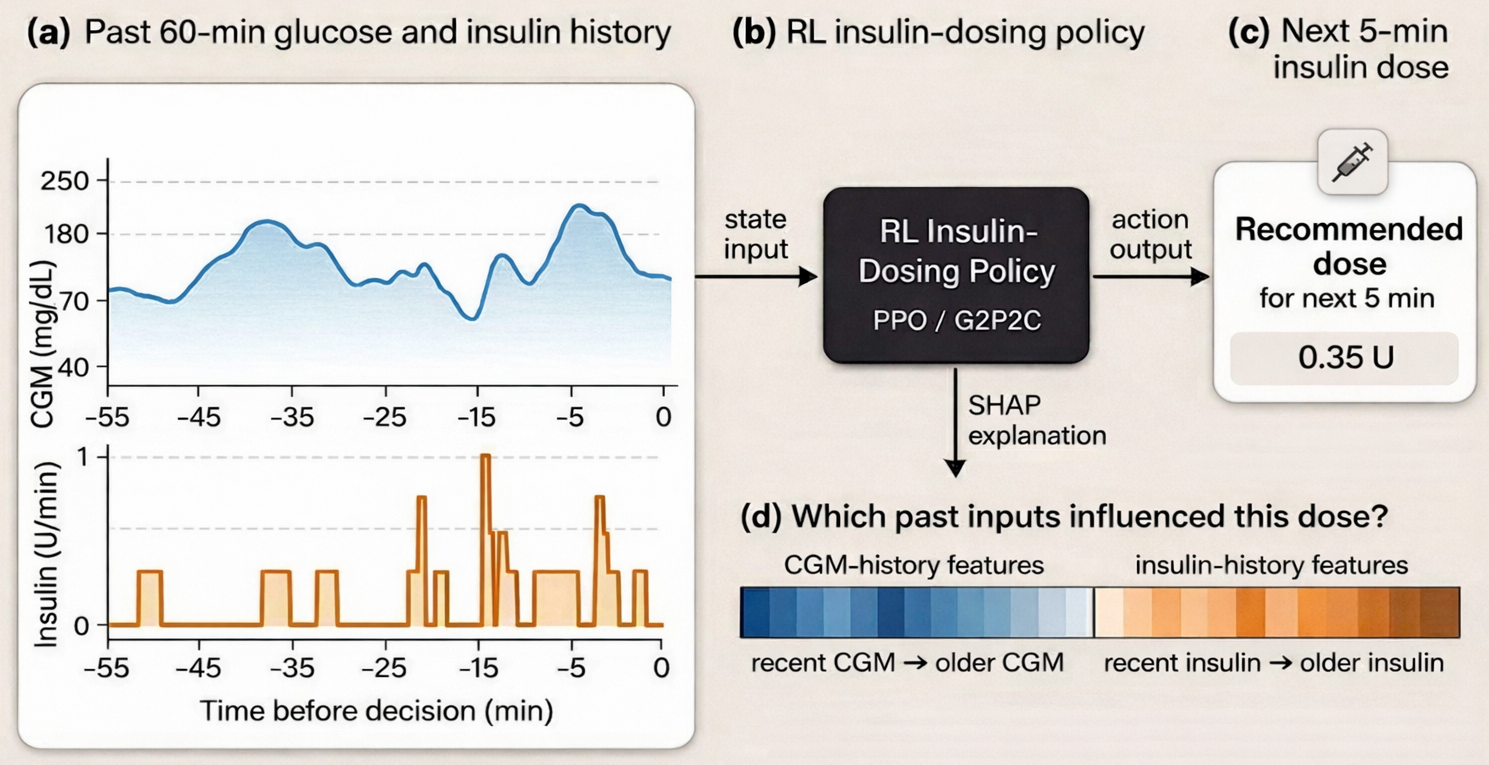}

    \caption{Overview of XRL4T1D. Every five minutes, an RL insulin-dosing policy receives a 60-minute history of CGM and administered-insulin measurements and produces the insulin recommendation for the next five-minute interval. SHAP attributes the recommendation to these historical inputs; insulin-history attributions are summarized over three 20-minute windows and checked using feature masking, direct window replacement, and TimeSHAP.}
    \label{fig:overview}
\end{figure*}

Both controllers rely predominantly on CGM history, particularly G2P2C. Within insulin history, the highest-ranked temporal window usually agrees with direct policy-output sensitivity, but the exact recent--middle--older profile and the PPO--G2P2C comparison vary across patient cohorts and explanation methods. In contrast, the overall feature rankings remain highly stable under the tested observation perturbations. Figure~\ref{fig:overview} summarizes the workflow.

Our contributions are threefold. 
\textbf{First, we develop and validate a post-hoc explanation framework for pretrained insulin-dosing policies.} We test whether inputs identified as important by SHAP produce larger changes in the policy output when replaced, both at the individual-feature level and after grouping insulin history into 20-min temporal windows.

\textbf{Second, we introduce a clinically motivated temporal summary of how each policy uses its insulin history.} The summary separates recent (0--20 min), middle (20--40 min), and older (40--60 min) portions of the observed hour and is interpreted together with the overall balance between CGM and insulin-history importance.

\textbf{Third, we evaluate how reliable these conclusions are across repeated simulations, glycemic contexts, observation perturbations, model checkpoints, and explanation settings.} Using PPO and G2P2C as case-study controllers, we find that individual-feature rankings and coarse temporal ordering are generally reliable, while finer controller comparisons can depend on patient cohort, explainer, and analysis settings. This distinction is important when using post-hoc explanations for preclinical policy comparison and monitoring.

\section{Related Work}
\label{sec:related}

RL insulin control has progressed from early and deep-RL formulations to recurrent, offline, safe-control, ensemble, and planning-based systems, with most development conducted in the UVA/Padova simulator \cite{ngo2018feedforward,fox2020deep,zhu2021jbhi,viroonluecha2022plos,emerson2023offline,visentin2014uvapadova,simglucose2018}. G2P2C is relevant here because it augments PPO with glucose prediction and short-horizon planning in a fully closed-loop setting \cite{hettiarachchi2024g2p2c}. This diversity makes outcome-only comparison incomplete: policies can reach similar aggregate metrics through different input sensitivities.

Explainability for sequential decision systems includes saliency, perturbation, attribution, policy summarization, and counterfactual approaches \cite{puiutta2020xrl,heuillet2020xrl,milani2024xrl,greydanus2018icml,juozapaitis2019ijcai,dong2025ijcai}. SHAP provides model-agnostic feature attributions \cite{lundberg2017shap}, while TimeSHAP adapts Shapley estimation to sequential inputs \cite{bento2021timeshap}. In a temporal attribution framework, however, cross-run stability or agreement between explainers does not alone show that highly ranked features affect the policy output; controlled masking is also needed.


Diabetes explainability has often focused on prediction rather than closed-loop control \cite{tasin2023xai,srinivasu2024xai,alredaini2026glucose}. Interpretability has also been explored in reinforcement-learning-based glucose control, including controller-specific designs incorporating attention or attribution \cite{lim2021access,adjevi2026xrl}. XRL4T1D differs primarily in its evaluation objective rather than its attribution algorithm: it examines pretrained policies without retraining and combines input-channel attribution with a clinically motivated temporal summary of administered-insulin history. Importantly, it evaluates whether the resulting feature and temporal-window rankings correspond to measured policy-output sensitivity under targeted replacement. This distinguishes the proposed workflow from reporting attribution magnitudes alone, while enabling systematic assessment of explanation stability across patients, scenarios, policy checkpoints, and explanation settings.

Recent work has also explored LLM-based insulin control by distilling an RL dosing policy into fine-tuned language models that reproduce dosing decisions and generate natural-language rationales \cite{sarkar2025explainable}. Our setting addresses a different question: XRL4T1D leaves the pretrained RL controller unchanged and tests whether post-hoc importance assigned to its physiological inputs corresponds to measurable policy-output sensitivity. This separates explanation of an existing controller from rationale generation by a language-model surrogate.
\section{Methodology}
\label{sec:methods}


\subsection{Experiment Setup and Evaluation Metrics}
\label{ssec:sim_env}

\paragraph{Simulation environment.}
We evaluate the policies in the FDA-accepted UVA/Padova Type~1 Diabetes simulator, accessed through \texttt{Simglucose} \cite{visentin2014uvapadova,simglucose2018}. The simulator models patient-specific glucose--insulin dynamics under exogenous insulin delivery and meal disturbances, enabling closed-loop evaluation across virtual subjects. We use 20 virtual patients (10 adolescents and 10 adults). Each evaluation is a 24-h closed-loop simulation comprising 288 five-minute decision steps. We consider one stochastic baseline meal protocol and two fixed meal protocols to evaluate the controllers under distinct but controlled meal disturbances; complete scenario and retrospective phase definitions are provided in Appendix~\ref{app:phase_results}. Meal timing affects the simulated glucose trajectory but is used only retrospectively for analysis and is not provided to the policies.

\paragraph{Reinforcement learning policies for glucose regulation.}
\label{ssec:policies}

We compare two publicly available pretrained controllers with a common input--output interface. PPO is a standard policy-gradient RL algorithm \cite{hettiarachchi2022reinforcement}, whereas G2P2C is a PPO-based controller that additionally incorporates glucose-prediction and short-horizon planning components \cite{hettiarachchi2024g2p2c}. Our comparison does not isolate the contribution of either added component.

At each 5-min decision time $t$, both policies map the state $s_t$ to a scalar insulin recommendation,
$\mu_t=\pi_\theta(s_t)\in\mathcal{A}$, where $\mathcal{A}\subset\mathbb{R}$ is the normalized action space \cite{hettiarachchi2022non}. The policy state contains 12 paired CGM and administered-insulin measurements spanning the preceding 60 min. In the stored flattened representation, the two channels are interleaved in temporal order:
\[
\mathbf{x}_t =
[
G_{t-55},I_{t-55},
G_{t-50},I_{t-50},
\ldots,
G_t,I_t
]
\in\mathbb{R}^{24}.
\]
Here, $G_{t-5j}$ and $I_{t-5j}$ denote the CGM and administered-insulin measurements $5j$ min before the current decision. Channel-wise analyses explicitly extract the corresponding CGM and insulin coordinates from this interleaved representation before temporal aggregation. We use the final trained patient-specific policies except in the checkpoint analysis.
\paragraph{Clinical outcome metrics.}
We report time in range (TIR; 70--180~mg/dL), time below range (TBR; below 70~mg/dL), time above range (TAR; above 180~mg/dL), and mean CGM using consensus thresholds \cite{battelino2019tir}. These outcomes describe glucose control, whereas the attribution analyses below characterize policy input dependence. A more comprehensive set of clinical outcome metrics is reported in Appendix~\ref{app:implementation}, Table~\ref{tab:clinical_baseline_full}.


\subsection{KernelSHAP Explanation}
\label{ssec:shap_framework}



We apply model-agnostic KernelSHAP to the policy output $\mu_t$ so that the same explanation procedure can be used for both controllers \cite{lundberg2017shap,strumbelj2014explaining}. KernelSHAP estimates the contribution of each input by comparing the current state with a set of reference states, referred to as the background set, while evaluating different subsets of input features, referred to as coalitions. We use 200 background states and 1,024 sampled coalitions by default. Each SHAP value $\phi_i$ measures the contribution assigned to feature $i$ relative to the empirical background distribution. Because adjacent CGM measurements and adjacent administered-insulin measurements are temporally correlated, these replacements can create input combinations that are unlikely under the observed data distribution. We therefore interpret SHAP values as post-hoc feature-importance scores rather than causal effects, and explicitly test whether their rankings correspond to policy-output sensitivity in Sec.~\ref{ssec:validation}.



\paragraph{Global and local explanations.}
Local SHAP values describe the feature contributions for an individual dosing decision. Global feature importance is computed as the mean absolute SHAP value over $N$ explained decisions:
\begin{equation}
    \Phi_i^{\mathrm{global}} = \frac{1}{N}\sum_{n=1}^{N} |\phi_i^{(n)}|.
    \label{eq:global_imp}
\end{equation}

\subsection{Clinically Inspired Temporal Insulin-Attribution Profile}
\label{ssec:iob_agg}

Insulin-on-board (IOB) is a clinical concept used in insulin therapy to describe insulin from previous doses that remains active in the body. Accounting for this residual insulin is important when determining subsequent insulin delivery, because delivering additional insulin while substantial prior insulin remains active can increase the risk of insulin stacking and hypoglycemia \cite{zisser2008bolus,walsh2014dia}. Motivated by this clinical view of recent versus more distal insulin exposure, we construct a temporal summary of how the policy uses its observed insulin history.

The policy observes 60 min of CGM and administered-insulin history, represented by 12 measurements from each channel sampled at 5-min intervals. We first quantify how the total absolute SHAP importance is divided between the two input channels. Let
\[
\Phi_{\mathrm{CGM}}
=
\sum_{j=0}^{11}
\left|\phi_{G_{t-5j}}\right|,
\qquad
\Phi_{\mathrm{INS}}
=
\sum_{j=0}^{11}
\left|\phi_{I_{t-5j}}\right|.
\]
We define the relative insulin-history importance as
\begin{equation}
r_{\mathrm{INS}}
=
\frac{\Phi_{\mathrm{INS}}}
{\Phi_{\mathrm{CGM}}+\Phi_{\mathrm{INS}}},
\label{eq:ins_reliance}
\end{equation}
with
\[
r_{\mathrm{CGM}} = 1-r_{\mathrm{INS}}.
\]
These quantities describe the relative importance assigned to the CGM and administered-insulin histories over the full 24-dimensional policy state.

We then examine how the insulin-history contribution is distributed over time. Rather than interpret the 12 insulin lags separately, we aggregate their absolute SHAP values into three non-overlapping 20-min windows: recent ($0$--$20$ min), middle ($20$--$40$ min), and older ($40$--$60$ min). We define
\[
\mathcal{K}_1=\{0,1,2,3\},\qquad
\mathcal{K}_2=\{4,5,6,7\},\qquad
\mathcal{K}_3=\{8,9,10,11\},
\]
corresponding to the recent, middle, and older windows, respectively. For $\Phi_{\mathrm{INS}}>0$, the normalized importance assigned to window $r$ is
\begin{equation}
w_r =
\frac{\sum_{j\in\mathcal{K}_r}
\left|\phi_{I_{t-5j}}\right|}
{\Phi_{\mathrm{INS}}}.
\label{eq:aiob}
\end{equation}
The three window fractions sum to one. Each $w_r$ therefore represents the fraction of \emph{insulin-history} importance assigned to that temporal window, rather than the fraction of total importance across all 24 policy inputs.

We refer to this clinically inspired temporal summary as \emph{Attribution-based Insulin-on-Board} (A-IOB). A-IOB is not a pharmacokinetic estimate of physiologically active insulin and should not be interpreted as conventional clinical IOB. Instead, it summarizes how the feature importance assigned to the policy's recorded insulin history is distributed across the preceding 60 min \cite{zisser2008bolus,walsh2014dia}. A-IOB therefore provides a clinically motivated temporal summary of the insulin-history inputs associated with the policy output.

For policy comparison, we report the CGM-versus-insulin importance split ($r_{\mathrm{CGM}}$ and $r_{\mathrm{INS}}$) together with the three-window A-IOB profile by cohort, glycemic phase, and meal scenario alongside conventional glycemic outcomes. Patient-level associations with insulin-to-carbohydrate ratio, insulin sensitivity factor, and total daily insulin are exploratory and are therefore reported separately in Appendix~\ref{app:e2}.

An important distinction is that the A-IOB fractions are conditional on the total attribution assigned to administered-insulin history. For example, a recent-window fraction of 0.6 indicates that 60\% of the insulin-history attribution falls within the most recent 20 minutes, not that this window contributes 60\% of the policy's total attribution or represents 60\% of physiologically active insulin. When overall insulin-history importance is small, normalized temporal fractions may exhibit substantial variation despite limited absolute attribution. We therefore interpret A-IOB jointly with the CGM-versus-insulin importance split, as illustrated in Fig. 
~\ref{fig:policy_profile}, and do not treat it as a pharmacokinetic estimate of insulin-on-board.

\subsection{Policy-Output and Explainer Agreement}
\label{ssec:validation}

\paragraph{Feature-level replacement.}
We first test whether the SHAP ranking of individual policy inputs reflects policy-output sensitivity. At each insulin-dosing decision, the policy state contains 24 scalar input features: 12 CGM measurements and 12 administered-insulin measurements from the preceding 60 min. We rank these 24 individual inputs by decreasing absolute attribution magnitude $|\phi_i|$.

For a mask size $m\in\{4,8\}$, we separately replace the $m$ inputs with the largest attribution magnitudes (\emph{top-$m$}) and the $m$ inputs with the smallest attribution magnitudes (\emph{bottom-$m$}) by their corresponding empirical background means, and measure the resulting absolute change in the normalized insulin recommendation. Thus, bottom-$m$ refers only to the $m$ least-attributed individual inputs, not to all inputs outside the top-$m$ set. As a descriptive reference, we also replace $m$ randomly selected individual inputs using a fixed random seed. If the feature-level attribution ranking is informative, replacing top-ranked inputs should produce a larger policy-output change than replacing bottom-ranked inputs.

Results are summarized over 80 patient--policy--mask-size cases
($20$ patients $\times$ $2$ policies $\times$ $2$ values of $m$); these cases are not treated as independent subjects.


\paragraph{Temporal-window replacement.}
We next test whether the three-window insulin-history summary reflects policy-output sensitivity. We evaluate 120 patient--policy--meal-scenario configurations ($20$ patients $\times$ $2$ policies $\times$ $3$ meal scenarios). For each configuration, we first extract the insulin-history coordinates from the interleaved policy state. We then replace the four insulin measurements belonging to one 20-min window at a time with their corresponding per-lag background means, while holding all CGM measurements and the remaining insulin-history measurements fixed. We compare the three windows by absolute policy-output change and by SHAP importance using Spearman $\rho$, and separately report whether the highest-ranked window is the same under both measures.

\paragraph{Sequential-explainer agreement.}
To test whether the temporal ordering depends on applying KernelSHAP to a flattened state representation, we additionally apply TimeSHAP \cite{bento2021timeshap} to the insulin-history channel for the same 120 configurations. The 12 insulin measurements are extracted from the interleaved policy state and arranged chronologically from oldest to newest to form the 12-event sequence, while the CGM inputs are held fixed. The reference is the mean insulin-history sequence from the corresponding 24-h evaluation simulation. TimeSHAP event scores are aggregated into the same recent, middle, and older windows, and their ordering is compared with KernelSHAP using Spearman $\rho$ and top-window agreement.

\subsection{Explanation Reproducibility and Sensitivity Analyses}
\label{ssec:reliability_methods}

\paragraph{Simulation stability.}
We repeat the stochastic baseline protocol across five random seeds to assess how the CGM-versus-insulin importance split and the within-insulin temporal profile vary across simulated trajectories.

\paragraph{Observation-corruption stability.}
We test explanation stability under two counterfactual CGM corruptions: additive Gaussian noise ($\sigma=10$ mg/dL) and $10\%$ randomly missing readings with forward-fill imputation. Clean observations continue to determine the policy action and simulated trajectory; the corrupted observations are maintained only in a separate history buffer used for explanation. For every comparison, the clean and corrupted histories are recorded at the same decision time before the subsequent environment transition, ensuring a paired comparison of the two explanations. This experiment therefore measures explanation stability rather than closed-loop controller robustness. For each patient and policy, we compute the 24-dimensional global feature-importance vector for the paired clean and corrupted histories using Eq.~\ref{eq:global_imp}, and compare their feature rankings using Spearman $\rho$.
\paragraph{Policy checkpoint sensitivity.}
To assess whether the framework can track changes across model versions, we evaluate all 20 patients per policy at episodes 50, 100, 150, and 195. For each checkpoint, we compute the 24-dimensional global feature-importance vector and compare its ranking with that of the final trained policy. We additionally track the three-window insulin-history profile across checkpoints.

\paragraph{KernelSHAP parameter sensitivity analysis.}
We evaluate eight representative patient--policy pairs, comprising four patients evaluated under both PPO and G2P2C. We vary the background-set size over $\{50,100,200,500\}$ and the number of sampled coalitions over $\{100,256,512,1024,2048\}$, yielding 20 explanation settings per pair. We examine how the resulting within-insulin temporal summaries vary across these settings. At the default setting, 95\% intervals are estimated using 200 bootstrap resamples \cite{efron1979}. Complete results are reported in Appendix~\ref{app:e4_protocol}, Table~\ref{tab:e4_full}.

\section{Results}
\label{sec:results}



\paragraph{Policy-Output Sensitivity of SHAP Rankings.}
\label{ssec:res_validation}

\begin{figure*}[h!]
  \centering
  \includegraphics[width=\textwidth]{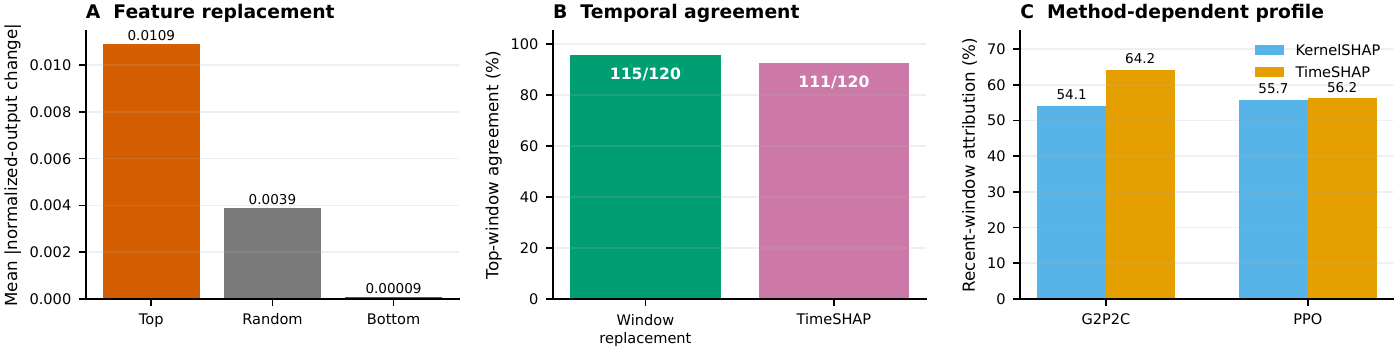}
  \caption{Policy-output validation of the temporal attribution framework. \textbf{A:} Feature-level replacement results pooled across PPO and G2P2C ($20$ patients $\times$ $2$ policies $\times$ $2$ mask sizes). Replacing the highest-attribution individual inputs produces the largest mean absolute change in the normalized insulin recommendation, while replacing the lowest-attribution inputs produces the smallest change. \textbf{B:} Temporal-window agreement pooled across PPO and G2P2C over $120$ patient--policy--meal-scenario configurations ($20$ patients $\times$ $2$ policies $\times$ $3$ scenarios). Bars show the fraction of configurations in which the highest-attribution window agrees with direct window replacement or TimeSHAP. \textbf{C:} Recent-window importance estimated by KernelSHAP and TimeSHAP for each controller. The methods give nearly identical values for PPO but differ more substantially for G2P2C, showing that agreement on the highest-ranked window does not imply agreement in the exact profile magnitude.
}
  \label{fig:validation_summary}
\end{figure*}

We first test the SHAP ranking at the individual-input level. At each decision, the 24 scalar policy inputs are ranked by absolute SHAP magnitude $|\phi_i|$. For a mask size $m\in\{4,8\}$, replacing the $m$ highest-ranked inputs produces a larger absolute change in the normalized insulin recommendation than replacing either the $m$ lowest-ranked or $m$ randomly selected inputs in all 80 patient--policy--mask-size cases ($20$ patients $\times$ $2$ policies $\times$ $2$ values of $m$). Pooled mean absolute output changes are 0.01090 for top-$m$, 0.00391 for random-$m$, and 0.000086 for bottom-$m$ replacement (Fig.~\ref{fig:validation_summary}A). We treat the 80/80 ordering as descriptive because each patient contributes both policies and both mask sizes.

This feature-level analysis is distinct from the temporal-window analysis below: here, we rank and replace individual scalar inputs, whereas the window-level analysis aggregates the 12 insulin-history attributions into three 20-min windows and compares the ranking of those three windows.

Two complementary checks evaluate the temporal summary. Across 120 patient--policy--meal-scenario configurations
($20$ patients $\times$ $2$ policies $\times$ $3$ meal scenarios), the highest-ranked insulin-history window under KernelSHAP is also the window whose replacement produces the largest policy-output change in 115/120 cases. Considering the ordering of all three windows, the mean Spearman correlation is $\rho=0.913$ (median $1.0$). TimeSHAP identifies the same highest-ranked window as KernelSHAP in 111/120 configurations, with mean three-window rank correlation $\rho=0.879$ (Fig.~\ref{fig:validation_summary}B--C). Thus, the coarse temporal ordering is generally reproducible, although neither comparison is exact and the resulting profile should not be interpreted as causal.

\paragraph{Policy Input-Dependence Profiles.}
\label{ssec:res_e1}

\begin{figure}[t]
  \centering
  \includegraphics[width=0.8\columnwidth]{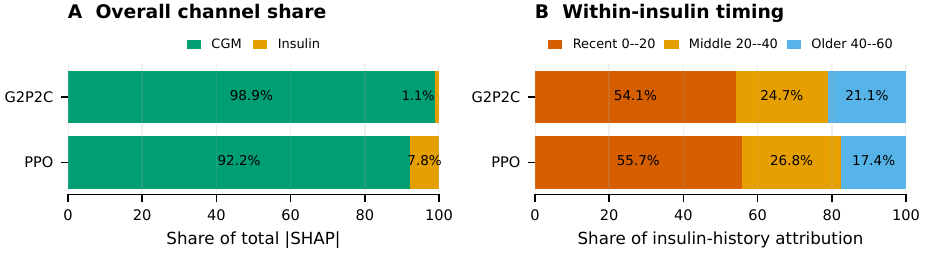}
  \caption{Policy input-dependence profiles across 120 patient--policy--meal-scenario configurations. \textbf{A:} Mean share of total absolute KernelSHAP importance assigned to CGM and administered-insulin history. Both controllers are CGM-dominated, particularly G2P2C. \textbf{B:} Mean distribution of insulin-history importance across recent (0--20 min), middle (20--40 min), and older (40--60 min) windows. Each temporal profile is normalized within the insulin-history channel before averaging.}
  \label{fig:policy_profile}
\end{figure}

The controller comparison is clearer at the input-channel level than in the normalized insulin-history profile. Both policies assign most of their total feature importance to CGM history, with this dependence particularly strong for G2P2C. Within the insulin-history channel, the pooled KernelSHAP profiles over the 120 patient--policy--meal-scenario configurations are similar: the recent window accounts for 54.15\% of insulin-history importance for G2P2C and 55.72\% for PPO, with the remaining importance distributed across the middle and older windows (Fig.~\ref{fig:policy_profile}). The temporal profile therefore does not support a general claim that one controller consistently places greater importance on recent insulin history.

The five-seed analysis shows that the within-insulin temporal profile varies more with cohort and phase than with controller identity: G2P2C has the larger recent-window fraction in only one of eight phase--cohort cells. The largest across-seed standard deviations are 5.11 percentage points for G2P2C and 6.26 for PPO. Full phase and cohort results are reported in Appendix Fig.~\ref{fig:e1_cgm_fraction} and Table~\ref{tab:e1_five_seed}.

Table~\ref{tab:clinical_baseline} places these explanation results beside baseline glycemic outcomes. TIR differs by 0.69 percentage points between policies in adolescents and 2.77 points in adults; no equivalence test was performed. Glycemic metrics summarize simulated glucose control, whereas the explanation analysis characterizes policy input dependence. Full outcomes are reported in Appendix Table~\ref{tab:clinical_baseline_full}.

\begin{table}[t]
\centering
\caption{Compact baseline outcomes across virtual patients, reported as median (IQR) per 24-h simulation. TIR/TBR/TAR are percentages; mean glucose is mg/dL.}
\label{tab:clinical_baseline}
\scriptsize
\resizebox{\columnwidth}{!}{%
\begin{tabular}{llcccc}
\toprule
Policy & Cohort & TIR & TBR & TAR & Mean glucose \\
\midrule
G2P2C & Adolescent & 58.16 (52.95--68.06) & 0.00 (0.00--0.17) & 40.45 (30.90--46.27) & 191.55 (163.43--217.75) \\
PPO   & Adolescent & 58.85 (51.22--67.97) & 0.00 (0.00--1.74) & 39.24 (31.51--47.31) & 187.33 (165.78--217.82) \\
G2P2C & Adult & 67.53 (60.50--73.09) & 0.00 (0.00--0.00) & 31.77 (25.69--38.11) & 166.68 (154.72--174.68) \\
PPO   & Adult & 64.76 (55.99--71.27) & 0.00 (0.00--0.17) & 34.03 (27.60--41.15) & 171.57 (159.08--180.89) \\
\bottomrule
\end{tabular}}
\end{table}

\paragraph{Model-version and explanation-parameter sensitivity.}
Across training checkpoints, the full feature ranking becomes progressively more similar to the final policy, whereas the recent-insulin fraction does not change monotonically (Appendix Fig.~\ref{fig:e4_drift}). KernelSHAP parameter sensitivity also varies across patients, with several G2P2C cases showing wider variation in the normalized within-insulin profile (Appendix Table~\ref{tab:e4_full}). These results show that stability of the global feature ranking does not necessarily imply stability of a derived temporal summary.

\section{Discussion}
\label{sec:discussion}

XRL4T1D checks post-hoc explanations against policy-output sensitivity before using them for controller comparison. Highly ranked individual inputs consistently produce larger output changes under replacement, and the highest-ranked insulin-history window usually matches the most output-sensitive window. TimeSHAP generally identifies the same highest-ranked window, although exact temporal proportions can differ. These results support feature-level and coarse temporal interpretation under the tested interventions, but do not show that SHAP recovers the internal mechanism of the policy.

\paragraph{Policy comparison and explanation stability.}
TIR and related glycemic metrics describe simulated glucose control, whereas XRL4T1D describes which recorded inputs are most strongly associated with the policy output. Both PPO and G2P2C rely predominantly on CGM history, particularly G2P2C, while their normalized within-insulin temporal profiles do not show a consistent controller-specific separation. Global feature rankings remain highly stable under the tested observation corruptions, but the within-insulin profile can vary across cohorts, explainers, and KernelSHAP settings. This distinction is important because A-IOB is normalized within the insulin channel: when insulin contributes little total importance, small changes within that channel can produce larger changes in the normalized temporal fractions.

\paragraph{Limitations.} Our evaluation is restricted to two pretrained RL controllers and 20 virtual patients in the UVA/Padova simulator. Although multiple meal scenarios, stochastic repetitions, policy checkpoints, and explanation configurations are considered, these analyses do not establish generalizability to other controller architectures, patient populations, or clinical settings. The simulations also exclude physiological disturbances such as exercise, illness, and infusion-site failures. The feature- and temporal-window replacement analyses use empirical background means, which may introduce implausible combinations of correlated observations and are not fully independent of the KernelSHAP reference mechanism. Consequently, agreement between attribution rankings and replacement-induced output changes should be interpreted as intervention-specific evidence rather than causal validation. Replacement strategies that preserve realistic temporal dependencies would provide a useful complementary assessment. Furthermore, A-IOB describes the relative distribution of attribution within the observed 60-minute insulin history and is not an estimate of physiological insulin-on-board. Its temporal fractions must be interpreted alongside total insulin-channel importance, particularly for policies with weak insulin-history attribution. Finally, the reported stability under observation corruption concerns explanations evaluated on paired clean trajectories, not closed-loop controller robustness. No clinician study or prospective clinical validation was conducted.

\section{Conclusion}
\label{sec:conclusion}

We presented XRL4T1D, a post-hoc explanation framework for pretrained RL insulin-dosing policies that tests SHAP-based feature importance against changes in the policy output. Across PPO and G2P2C controllers evaluated on 20 virtual patients, highly ranked individual inputs consistently produced larger output changes under replacement, and the highest-ranked insulin-history window agreed with direct window replacement in most configurations. Both controllers relied predominantly on CGM history, while the finer recent--middle--older insulin profile varied across cohorts, explanation methods, and analysis settings. At the same time, full feature rankings remained highly stable under the tested observation perturbations and could be tracked across policy checkpoints. These results show that validated post-hoc explanation can provide a reproducible basis for comparing RL insulin-dosing policies across clinically relevant glycemic phases, meal scenarios, and sensor perturbations, supporting benchmark-style preclinical evaluation and monitoring of how autonomous controllers use longitudinal physiological information. These findings remain descriptive rather than causal and do not establish clinical superiority of either controller or attribute the observed differences to a specific architectural component.


\bibliographystyle{plainnat}
\bibliography{references}

\appendix
\section{Additional Clinical Outcomes and Implementation Details}
\label{app:implementation}

\subsection{Patient descriptors}
The insulin-to-carbohydrate ratio (ICR) is the grams of carbohydrate covered by one unit of rapid-acting insulin; the insulin sensitivity factor (ISF) is the expected glucose reduction from one unit; and total daily insulin (TDI) is insulin delivered over 24 hours \cite{schiavon2018cr,davidson2008basalbolus,janez2020insulin}. These descriptors are used only in the exploratory patient-level analyses in Appendix~\ref{app:e2}.

\subsection{Complete baseline glycemic outcomes}
Table~\ref{tab:clinical_baseline_full} expands the baseline outcomes in Table~\ref{tab:clinical_baseline}. Each policy--cohort row aggregates 10 patient-specific 24-h simulations. Similar descriptive values do not establish statistical equivalence; no equivalence test was performed.

\begin{table*}[t]
\centering
\caption{Complete baseline glycemic outcomes. Continuous metrics are median (IQR) across 10 patient-specific 24-h simulations per policy--cohort cell; event and swing counts are mean $\pm$ SD per simulation. TIR, TBR, TAR, and CV are percentages, and mean glucose is in mg/dL. Severe episodes use thresholds below 54 or above 250~mg/dL.}
\label{tab:clinical_baseline_full}
\scriptsize
\resizebox{\textwidth}{!}{%
\begin{tabular}{llcccccccc}
\toprule
Policy & Cohort & TIR & TBR & TAR & Mean glucose & CV & Hypo / severe & Hyper / severe & Swings \\
\midrule
G2P2C & Adolescent & 58.16 (52.95--68.06) & 0.00 (0.00--0.17) & 40.45 (30.90--46.27) & 191.55 (163.43--217.75) & 49.22 (42.26--57.33) & 0.45$\pm$0.96 / 0.05$\pm$0.28 & 2.72$\pm$0.82 / 2.27$\pm$0.95 & 49.22$\pm$26.57 \\
PPO   & Adolescent & 58.85 (51.22--67.97) & 0.00 (0.00--1.74) & 39.24 (31.51--47.31) & 187.33 (165.78--217.82) & 49.88 (43.60--56.85) & 0.60$\pm$0.95 / 0.05$\pm$0.22 & 2.67$\pm$0.81 / 2.15$\pm$0.93 & 47.75$\pm$27.35 \\
G2P2C & Adult & 67.53 (60.50--73.09) & 0.00 (0.00--0.00) & 31.77 (25.69--38.11) & 166.68 (154.72--174.68) & 36.62 (28.21--41.99) & 0.40$\pm$1.02 / 0.07$\pm$0.40 & 3.10$\pm$0.89 / 1.63$\pm$0.82 & 23.03$\pm$12.66 \\
PPO   & Adult & 64.76 (55.99--71.27) & 0.00 (0.00--0.17) & 34.03 (27.60--41.15) & 171.57 (159.08--180.89) & 36.55 (29.02--41.30) & 0.52$\pm$1.06 / 0.15$\pm$0.60 & 2.88$\pm$0.75 / 1.83$\pm$0.78 & 23.20$\pm$12.99 \\
\bottomrule
\end{tabular}}
\end{table*}

\subsection{Implementation and runtime}
The default KernelSHAP configuration uses a background set of 200 states and 1,024 sampled coalitions per explanation. On a Windows~11 workstation with a 16-core Intel processor, 16~GB RAM, and an NVIDIA RTX~4060 Laptop GPU, a 10-decision benchmark took 2.15~s (0.215~s per explanation). Policy forward passes used the GPU and the KernelSHAP solver used the CPU. This hardware-specific offline explanation time is not a measurement of clinical latency or deployment readiness.

\FloatBarrier
\section{Full Evaluation Setting, Phase, Cohort, and Meal-Scenario Results}
\label{app:phase_results}

\subsection{Meal scenarios and retrospective glycemic phases}
Scenario~1 is stochastic: breakfast, lunch, and dinner are each enabled with probability 0.95 and use nominal carbohydrate amounts of 45, 85, and 80~g, with configured amount-variation parameters of 5, 5, and 10~g and timing-variation parameters of 60~min around the nominal 08{:}00, 13{:}00, and 20{:}00 meal times. Scenario~2 uses deterministic 40/80/60~g meals at 08{:}00/13{:}00/20{:}00, and Scenario~3 uses deterministic 45/85/80~g meals at those times. The same pretrained policies are evaluated without retraining.

Each 24-h simulation is partitioned retrospectively into preprandial ($-30$ to 0~min), early postprandial (0 to $+60$~min), late postprandial ($+60$ to $+180$~min), and nocturnal (00{:}00--06{:}00) strata \cite{slattery2018timing,lupoli2019postprandial,bell2015postprandial,wu2022nocturnal}. Meal timing is used only to assign these analysis labels and is absent from the 24-dimensional policy input.

\subsection{Overall CGM-versus-insulin importance}
\begin{figure*}[t]
  \centering
  \includegraphics[width=\textwidth]{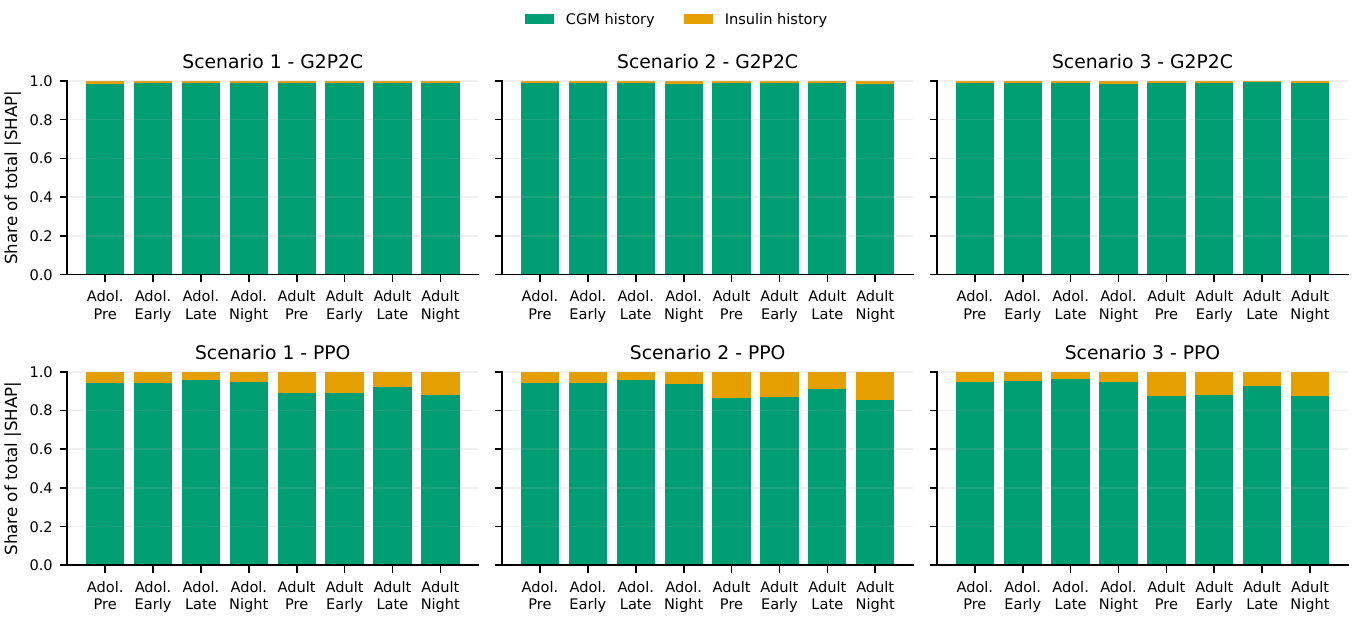}
  \caption{CGM-versus-insulin importance across 20 patients, two policies, three meal scenarios, and four retrospective phases. Each bar is the mean of 10 patient-level total-$|\mathrm{SHAP}|$ channel shares for the indicated cohort--policy--scenario--phase cell; green is $r_{\mathrm{CGM}}$ and orange is $r_{\mathrm{INS}}$. Adol. denotes the adolescent cohort.}
  \label{fig:e1_cgm_fraction}
\end{figure*}

The stored state is interleaved from oldest to newest,
\[
[G_{t-55},I_{t-55},G_{t-50},I_{t-50},\ldots,G_t,I_t].
\]
Channel-specific analyses explicitly extract the corresponding CGM and administered-insulin entries from this interleaved representation before aggregation.
For each decision,
\[
\Phi_{\mathrm{CGM}}=\sum_{j=0}^{11}|\phi_{G_{t-5j}}|,
\qquad
\Phi_{\mathrm{INS}}=\sum_{j=0}^{11}|\phi_{I_{t-5j}}|,
\]
and
\[
r_{\mathrm{INS}}=\frac{\Phi_{\mathrm{INS}}}
{\Phi_{\mathrm{CGM}}+\Phi_{\mathrm{INS}}},
\qquad r_{\mathrm{CGM}}=1-r_{\mathrm{INS}}.
\]
Figure~\ref{fig:e1_cgm_fraction} reports these total-channel shares for every scenario, cohort, phase, and policy cell. Both policies are CGM-dominated, with the stronger CGM dominance generally observed for G2P2C.

\begin{table*}[p]
\centering
\caption{Exact total-channel shares corresponding to Fig.~\ref{fig:e1_cgm_fraction}. Entries are percentages averaged across 10 patients within each cohort--policy--scenario--phase cell. CGM and administered-insulin shares use total absolute KernelSHAP importance as their common denominator.}
\label{tab:e1_channel_cells}
\scriptsize
\begin{tabular}{cllrrrr}
\toprule
Scenario & Policy & Phase & Adol. CGM & Adol. insulin & Adult CGM & Adult insulin \\
\midrule
1 & G2P2C & Pre   & 98.66 & 1.34 & 98.95 & 1.05 \\
1 & G2P2C & Early & 98.85 & 1.15 & 98.92 & 1.08 \\
1 & G2P2C & Late  & 98.94 & 1.06 & 99.17 & 0.83 \\
1 & G2P2C & Night & 98.75 & 1.25 & 98.79 & 1.21 \\
1 & PPO   & Pre   & 94.02 & 5.98 & 88.88 & 11.12 \\
1 & PPO   & Early & 94.07 & 5.93 & 89.24 & 10.76 \\
1 & PPO   & Late  & 95.88 & 4.12 & 92.17 & 7.83 \\
1 & PPO   & Night & 94.50 & 5.50 & 88.08 & 11.92 \\
\addlinespace
2 & G2P2C & Pre   & 98.70 & 1.30 & 98.73 & 1.27 \\
2 & G2P2C & Early & 98.81 & 1.19 & 98.84 & 1.16 \\
2 & G2P2C & Late  & 98.96 & 1.04 & 99.16 & 0.84 \\
2 & G2P2C & Night & 98.66 & 1.34 & 98.56 & 1.44 \\
2 & PPO   & Pre   & 94.13 & 5.87 & 86.53 & 13.47 \\
2 & PPO   & Early & 94.25 & 5.75 & 86.94 & 13.06 \\
2 & PPO   & Late  & 95.60 & 4.40 & 91.31 & 8.69 \\
2 & PPO   & Night & 93.76 & 6.24 & 85.25 & 14.75 \\
\addlinespace
3 & G2P2C & Pre   & 98.70 & 1.30 & 98.86 & 1.14 \\
3 & G2P2C & Early & 98.83 & 1.17 & 98.94 & 1.06 \\
3 & G2P2C & Late  & 98.98 & 1.02 & 99.21 & 0.79 \\
3 & G2P2C & Night & 98.69 & 1.31 & 98.74 & 1.26 \\
3 & PPO   & Pre   & 94.83 & 5.17 & 87.67 & 12.33 \\
3 & PPO   & Early & 95.00 & 5.00 & 88.01 & 11.99 \\
3 & PPO   & Late  & 96.23 & 3.77 & 92.46 & 7.54 \\
3 & PPO   & Night & 94.53 & 5.47 & 87.25 & 12.75 \\
\bottomrule
\end{tabular}
\end{table*}

\subsection{Within-insulin A-IOB profile}
Conditional on insulin-history importance, the A-IOB fractions are
\[
w_r=\frac{\sum_{j\in\mathcal K_r}|\phi_{I_{t-5j}}|}{\Phi_{\mathrm{INS}}},
\quad r\in\{0\text{--}20,20\text{--}40,40\text{--}60\}.
\]
Thus, $w_{0\text{--}20}=0.55$ means that 55\% of the importance assigned to administered-insulin history lies in the recent window; it does not mean that recent insulin accounts for 55\% of total model importance. Tables~\ref{tab:e1_iob_phase}--\ref{tab:e1_iob_phase_hard} should therefore be interpreted alongside Fig.~\ref{fig:e1_cgm_fraction}. The tables do not support a universal G2P2C recent-window advantage.

\begin{table*}[p]
\centering
\caption{Scenario~1 within-insulin A-IOB profiles. Entries are means across 10 patients in each cohort--policy--phase cell from one stochastic-baseline 24-h simulation per patient. Fractions are normalized within administered-insulin history and should be interpreted with the total insulin shares in Fig.~\ref{fig:e1_cgm_fraction}.}
\label{tab:e1_iob_phase}
\scriptsize
\begin{tabular}{lllccc@{\qquad}lllccc}
\toprule
Cohort & Policy & Phase & Recent & Middle & Older & Cohort & Policy & Phase & Recent & Middle & Older \\
\midrule
Adol. & G2P2C & Pre   & .439 & .320 & .241 & Adult & G2P2C & Pre   & .565 & .243 & .193 \\
Adol. & G2P2C & Early & .437 & .285 & .278 & Adult & G2P2C & Early & .562 & .248 & .190 \\
Adol. & G2P2C & Late  & .445 & .292 & .263 & Adult & G2P2C & Late  & .577 & .219 & .205 \\
Adol. & G2P2C & Night & .485 & .281 & .234 & Adult & G2P2C & Night & .626 & .213 & .162 \\
\addlinespace
Adol. & PPO & Pre   & .475 & .284 & .241 & Adult & PPO & Pre   & .623 & .238 & .139 \\
Adol. & PPO & Early & .490 & .266 & .243 & Adult & PPO & Early & .628 & .245 & .128 \\
Adol. & PPO & Late  & .501 & .276 & .223 & Adult & PPO & Late  & .626 & .249 & .126 \\
Adol. & PPO & Night & .511 & .280 & .209 & Adult & PPO & Night & .619 & .249 & .133 \\
\bottomrule
\end{tabular}
\end{table*}

\begin{table*}[p]
\centering
\caption{Scenario~2 within-insulin A-IOB profiles. Entries are means across 10 patients per cohort--policy--phase cell. Fractions are normalized within administered-insulin history and should be interpreted with Fig.~\ref{fig:e1_cgm_fraction}.}
\label{tab:e1_iob_phase_easy}
\scriptsize
\begin{tabular}{lllccc@{\qquad}lllccc}
\toprule
Cohort & Policy & Phase & Recent & Middle & Older & Cohort & Policy & Phase & Recent & Middle & Older \\
\midrule
Adol. & G2P2C & Pre   & .440 & .297 & .263 & Adult & G2P2C & Pre   & .612 & .230 & .158 \\
Adol. & G2P2C & Early & .465 & .307 & .228 & Adult & G2P2C & Early & .596 & .228 & .177 \\
Adol. & G2P2C & Late  & .474 & .278 & .248 & Adult & G2P2C & Late  & .602 & .218 & .180 \\
Adol. & G2P2C & Night & .496 & .272 & .233 & Adult & G2P2C & Night & .650 & .197 & .153 \\
\addlinespace
Adol. & PPO & Pre   & .517 & .272 & .211 & Adult & PPO & Pre   & .620 & .252 & .128 \\
Adol. & PPO & Early & .501 & .280 & .219 & Adult & PPO & Early & .603 & .249 & .148 \\
Adol. & PPO & Late  & .513 & .271 & .216 & Adult & PPO & Late  & .628 & .243 & .129 \\
Adol. & PPO & Night & .538 & .270 & .192 & Adult & PPO & Night & .615 & .250 & .135 \\
\bottomrule
\end{tabular}
\end{table*}

\begin{table*}[p]
\centering
\caption{Scenario~3 within-insulin A-IOB profiles. Entries are means across 10 patients per cohort--policy--phase cell. Fractions are normalized within administered-insulin history and should be interpreted with Fig.~\ref{fig:e1_cgm_fraction}.}
\label{tab:e1_iob_phase_hard}
\scriptsize
\begin{tabular}{lllccc@{\qquad}lllccc}
\toprule
Cohort & Policy & Phase & Recent & Middle & Older & Cohort & Policy & Phase & Recent & Middle & Older \\
\midrule
Adol. & G2P2C & Pre   & .416 & .307 & .276 & Adult & G2P2C & Pre   & .614 & .220 & .165 \\
Adol. & G2P2C & Early & .453 & .293 & .254 & Adult & G2P2C & Early & .587 & .217 & .196 \\
Adol. & G2P2C & Late  & .453 & .268 & .279 & Adult & G2P2C & Late  & .558 & .237 & .205 \\
Adol. & G2P2C & Night & .498 & .269 & .232 & Adult & G2P2C & Night & .638 & .207 & .155 \\
\addlinespace
Adol. & PPO & Pre   & .517 & .275 & .208 & Adult & PPO & Pre   & .620 & .253 & .127 \\
Adol. & PPO & Early & .530 & .272 & .198 & Adult & PPO & Early & .602 & .250 & .149 \\
Adol. & PPO & Late  & .516 & .273 & .211 & Adult & PPO & Late  & .630 & .244 & .126 \\
Adol. & PPO & Night & .540 & .270 & .190 & Adult & PPO & Night & .614 & .250 & .135 \\
\bottomrule
\end{tabular}
\end{table*}

\subsection{Five-seed reproducibility}
Only Scenario~1 is repeated across stochastic seeds; Scenarios~2 and 3 are deterministic and are excluded from this analysis. Table~\ref{tab:e1_five_seed} reports mean $\pm$ SD across five seed-level cohort summaries. The recent fraction varies by cohort and phase, and G2P2C exceeds PPO in only one of the eight matched phase--cohort cells. The largest recent-fraction SD is 0.051 for G2P2C and 0.063 for PPO.

\begin{table*}[t]
\centering
\caption{Five-seed Scenario~1 reproducibility. Each entry is mean $\pm$ SD across five seed-level summaries after aggregating patients within the indicated cohort, policy, and phase. Recent is $w_{0\text{--}20}$ normalized within administered-insulin history; CGM and insulin are percentages of total absolute attribution.}
\label{tab:e1_five_seed}
\scriptsize
\begin{tabular}{lllccc@{\qquad}lllccc}
\toprule
Cohort & Phase & Policy & Recent & CGM (\%) & Ins. (\%) & Cohort & Phase & Policy & Recent & CGM (\%) & Ins. (\%) \\
\midrule
Adol. & Pre   & G2P2C & .501$\pm$.051 & 99.45$\pm$.03 & .55$\pm$.03 & Adult & Pre   & G2P2C & .660$\pm$.046 & 99.02$\pm$.12 & .98$\pm$.12 \\
Adol. & Early & G2P2C & .461$\pm$.042 & 99.31$\pm$.18 & .69$\pm$.18 & Adult & Early & G2P2C & .612$\pm$.045 & 98.90$\pm$.20 & 1.10$\pm$.20 \\
Adol. & Late  & G2P2C & .468$\pm$.048 & 99.52$\pm$.05 & .48$\pm$.05 & Adult & Late  & G2P2C & .610$\pm$.051 & 99.21$\pm$.11 & .79$\pm$.11 \\
Adol. & Night & G2P2C & .491$\pm$.021 & 99.42$\pm$.12 & .58$\pm$.12 & Adult & Night & G2P2C & .671$\pm$.030 & 98.87$\pm$.35 & 1.13$\pm$.35 \\
\addlinespace
Adol. & Pre   & PPO & .569$\pm$.063 & 96.28$\pm$.35 & 3.72$\pm$.35 & Adult & Pre   & PPO & .671$\pm$.060 & 91.41$\pm$.50 & 8.59$\pm$.50 \\
Adol. & Early & PPO & .499$\pm$.036 & 96.42$\pm$.94 & 3.58$\pm$.94 & Adult & Early & PPO & .618$\pm$.033 & 91.38$\pm$2.27 & 8.62$\pm$2.27 \\
Adol. & Late  & PPO & .517$\pm$.049 & 97.72$\pm$.39 & 2.28$\pm$.39 & Adult & Late  & PPO & .648$\pm$.038 & 94.88$\pm$.98 & 5.12$\pm$.98 \\
Adol. & Night & PPO & .533$\pm$.039 & 96.03$\pm$1.13 & 3.97$\pm$1.13 & Adult & Night & PPO & .630$\pm$.024 & 90.05$\pm$2.68 & 9.95$\pm$2.68 \\
\bottomrule
\end{tabular}
\end{table*}

\FloatBarrier
\section{Exploratory Patient-Level Clinical Alignment}
\label{app:e2}

\subsection{Setup}
\label{app:e2_setup}
The exploratory analysis uses the 0--120~min postprandial windows of Scenario~1 for 20 patients (10 adolescents and 10 adults), analyzed separately for PPO and G2P2C. Patient descriptors are TDI, ISF, and ICR. Cohorts are pooled for the correlations below. Because the sample is small and multiple associations are examined, all results are descriptive and hypothesis-generating.

\subsection{Patient-level heatmaps}
\label{app:e2_heatmaps}
\begin{figure*}[t]
  \centering
  \begin{subfigure}{0.9\textwidth}
    \centering
    \includegraphics[width=\linewidth]{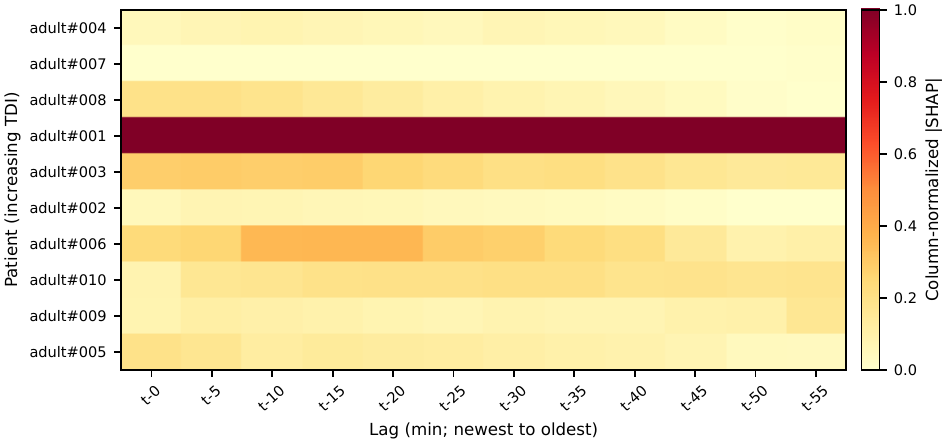}
    \caption{Administered-insulin history.}
  \end{subfigure}

  \vspace{0.5em}

  \begin{subfigure}{0.9\textwidth}
    \centering
    \includegraphics[width=\linewidth]{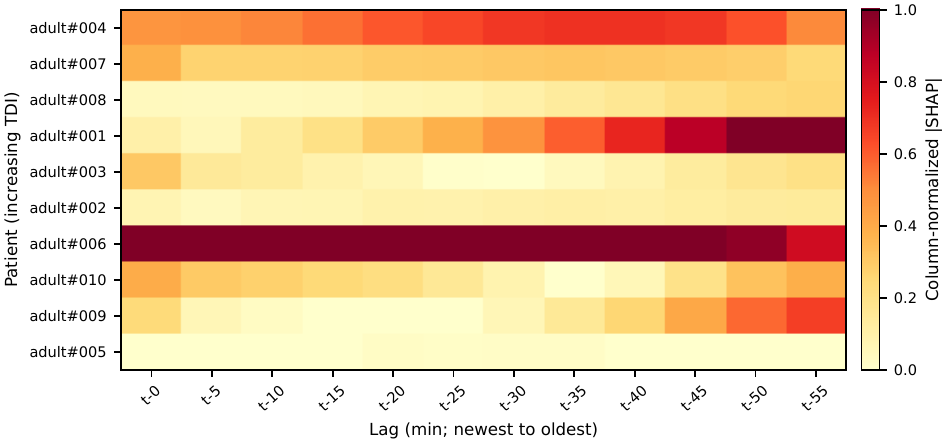}
    \caption{CGM history.}
  \end{subfigure}
  \caption{Postprandial mean absolute KernelSHAP importance for 10 adult PPO patients in Scenario~1, sorted by increasing TDI. Columns run from the newest to the oldest lag and are normalized separately across patients. The panels show temporal variation within each channel, not cross-channel magnitude or physiological effects.}
  \label{fig:e2_heatmaps}
\end{figure*}

Figure~\ref{fig:e2_heatmaps} shows the 10 adult PPO patients, sorted by TDI. CGM and administered-insulin coordinates are extracted from the interleaved state before plotting. Each lag column is scaled separately to $[0,1]$ across patients, so color intensity compares patients within a lag and must not be used to compare absolute magnitude across channels.

\subsection{Channel and temporal summaries}
\label{app:e2_cgm}
Table~\ref{tab:e2_channel_timing} reports the pooled postprandial summaries. Overall insulin share and the A-IOB fractions have different denominators: $r_{\mathrm{INS}}$ is relative to both input channels, whereas recent/middle/older fractions are conditional on insulin-history importance.

\begin{table}[t]
\centering
\caption{Scenario~1 postprandial summaries across 20 patients per policy. Values are patient means. CGM and insulin are percentages of total absolute attribution; the three temporal fractions are normalized within administered-insulin history.}
\label{tab:e2_channel_timing}
\small
\begin{tabular}{lccccc}
\toprule
Policy & CGM (\%) & Ins. (\%) & Recent & Middle & Older \\
\midrule
G2P2C & 98.99 & 1.01 & .527 & .256 & .217 \\
PPO   & 92.34 & 7.66 & .560 & .262 & .178 \\
\bottomrule
\end{tabular}
\end{table}

\subsection{Exploratory associations}
\label{app:e2_align}
Table~\ref{tab:e2_correlations} lists the six smallest uncorrected $p$-values among 60 policy-specific descriptor--attribution comparisons. The most consistent pattern is a negative association between ISF and overall insulin share for both policies. These associations do not show adaptation, clinical benefit, or causality.

\begin{table}[t]
\centering
\caption{Exploratory Spearman associations across the pooled adolescent and adult cohorts ($n=20$ per row). Insulin share is $r_{\mathrm{INS}}$. Values are uncorrected; 60 comparisons were examined and no multiplicity correction was applied.}
\label{tab:e2_correlations}
\small
\begin{tabular}{lllrr}
\toprule
Policy & Patient descriptor & Attribution summary & $\rho$ & $p$ \\
\midrule
PPO   & ISF & Insulin share & -.690 & .0008 \\
G2P2C & ISF & Insulin share & -.680 & .0010 \\
G2P2C & TDI & Insulin share &  .526 & .0171 \\
G2P2C & ICR & Insulin share & -.516 & .0200 \\
PPO   & ICR & Insulin share & -.449 & .0471 \\
PPO   & TDI & Insulin share &  .444 & .0501 \\
\bottomrule
\end{tabular}
\end{table}

The small sample limits precision, and pooling cohorts can confound associations. The results are hypothesis-generating and should not be interpreted as patient-specific policy adaptation or physiological validation of A-IOB.

\FloatBarrier
\section{Reliability and Sensitivity Analyses}
\label{app:e2e4}

\subsection{Observation-corruption protocol}
\label{app:e2_protocol}
The policy action and simulated trajectory are always generated from the clean observation history. For explanation only, a separate buffer applies either Gaussian CGM noise, $\widetilde G_t=G_t+\epsilon_t$ with $\epsilon_t\sim\mathcal N(0,10^2)$~mg/dL, or 10\% missing readings with forward-fill, $\widetilde G_t=\widetilde G_{t-1}$. Clean and corrupted histories are paired at the same decision time. This tests explanation stability, not closed-loop controller robustness to sensor failure.

\subsection{Observation-corruption results}
\label{app:e2_results}
Table~\ref{tab:e2_stats} summarizes the 80 patient--policy--corruption comparisons. Every rank correlation exceeds 0.926 and no case falls below the prespecified 0.85 threshold. Top-10 feature overlap is also high. These values describe the stability of global attribution rankings on a fixed clean trajectory.

\begin{figure}[t]
    \centering
    \includegraphics[width=0.95\columnwidth]{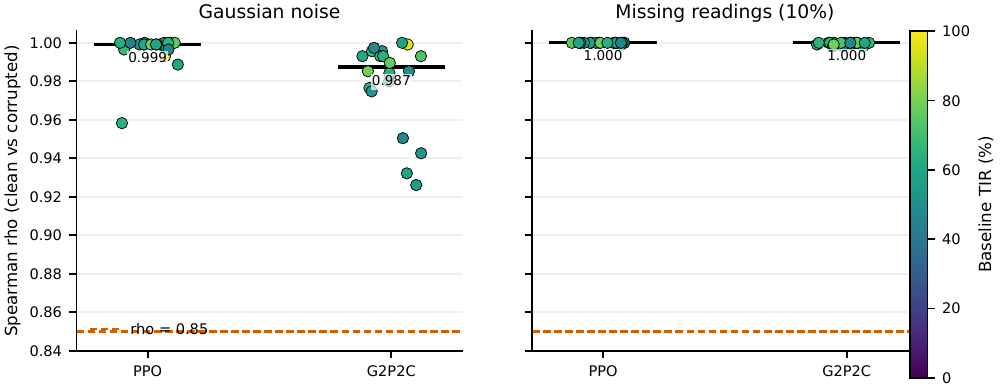}
    \caption{Stability of per-patient global feature rankings under counterfactual observation corruption. Spearman $\rho$ compares clean and corrupted global feature-importance vectors recorded at the same decision time; color shows baseline TIR only as a visual reference. Rankings remain highly stable for both policies: all 80 comparisons have $\rho>0.926$, and none falls below the $\rho=0.85$ reference threshold.}
    \label{fig:e3_strip}
\end{figure}

The global feature ranking is highly stable under both tested observation corruptions. Under Gaussian CGM noise, the median clean-versus-corrupted Spearman correlation is 0.999 for PPO and 0.987 for G2P2C; under 10\% missing readings, the median is 1.000 for both policies. Across all 80 patient--policy--corruption ($20$ patients $\times$ $2$ policies $\times$ $2$ corruption modes) comparisons, the minimum correlation is 0.926 and no case falls below $\rho=0.85$. This experiment measures explanation stability for paired histories on the same clean action trajectory, not controller safety under sensor failure.

\begin{table}[t]
\centering
\caption{Explanation stability under counterfactual observation corruption. Each row contains 20 paired clean/corrupted patient comparisons. $\rho$ is the Spearman correlation between 24-feature global-importance vectors; overlap is the mean fraction of shared top-10 features.}
\label{tab:e2_stats}
\small
\begin{tabular}{llrrrr}
\toprule
Policy & Corruption & Median $\rho$ & Min. $\rho$ & $\rho<.85$ & Top-10 overlap \\
\midrule
PPO   & Gaussian CGM noise & .999 & .958 & 0/20 & 1.00 \\
G2P2C & Gaussian CGM noise & .987 & .926 & 0/20 & .98 \\
PPO   & 10\% missing + fill & 1.000 & .999 & 0/20 & 1.00 \\
G2P2C & 10\% missing + fill & 1.000 & .998 & 0/20 & 1.00 \\
\bottomrule
\end{tabular}
\end{table}

\subsection{Policy-checkpoint sensitivity}
\begin{figure*}[t]
  \centering
  \includegraphics[width=0.82\textwidth]{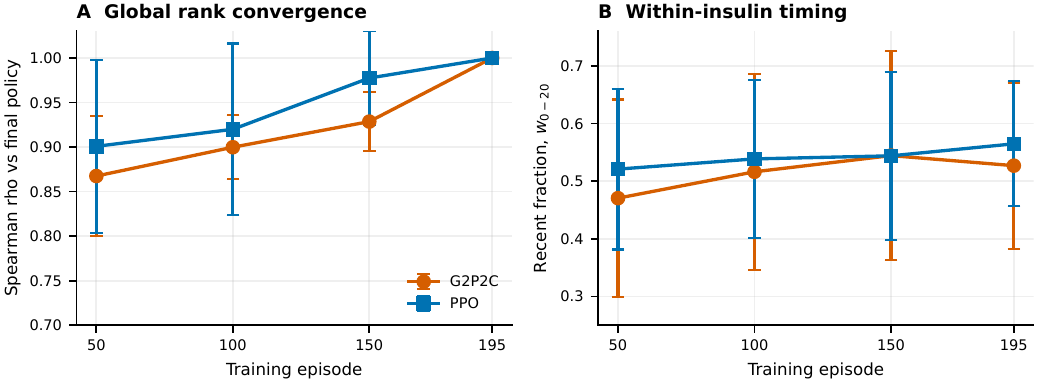}
  \caption{Checkpoint sensitivity across 20 patients per policy. \textbf{A:} Mean Spearman correlation of the full 24-feature ranking at each earlier checkpoint with the episode-195 policy; the final checkpoint is 1 by definition. \textbf{B:} Mean A-IOB recent fraction $w_{0\text{--}20}$ at each checkpoint. Error bars are inter-patient SD. Global rankings approach the final policy, while the G2P2C recent fraction is non-monotonic.}
  \label{fig:e4_drift}
\end{figure*}

Figure~\ref{fig:e4_drift} treats checkpoints as model versions. Earlier-to-final global feature-rank correlations rise from 0.867/0.901 at episode~50 to 0.929/0.977 at episode~150 for G2P2C/PPO. In contrast, G2P2C's mean within-insulin recent fraction is non-monotonic (0.471, 0.516, 0.544, and 0.527 at episodes 50, 100, 150, and 195); PPO changes from 0.521 to 0.565 over the same checkpoints. The comparison tracks model versions and does not isolate an architectural cause.

\subsection{KernelSHAP parameter sensitivity}
\label{app:e4_protocol}
The 20-setting sweep crosses background sizes $N_{\mathrm{bg}}\in\{50,100,200,500\}$ with coalition counts $N_{\mathrm{samples}}\in\{100,256,512,1024,2048\}$ for eight patient--policy pairs. Table~\ref{tab:e4_full} shows heterogeneous within-insulin sensitivity: three G2P2C cases span more than 0.31 in recent fraction, whereas other cases are much narrower. This is compatible with stable global rankings but shows that a normalized within-insulin profile can be sensitive when total insulin importance is small.

At the default $(N_{\mathrm{bg}},N_{\mathrm{samples}})=(200,1024)$ setting, each 95\% interval is based on 200 bootstrap resamples of explanation timesteps with replacement within that patient. The stored reference-rank field is populated for only 7 of 20 settings per patient, so no median reference correlation is reported.

\begin{table*}[t]
\centering
\caption{KernelSHAP parameter sensitivity for eight patient--policy pairs. Profile and look-back ranges cover all 20 settings. $t_{50}$ and $t_{80}$ are minutes. The final column is the default-setting bootstrap 95\% interval for $w_{0\text{--}20}$; all temporal fractions are normalized within administered-insulin history.}
\label{tab:e4_full}
\small
\begin{tabular}{llcccc}
\toprule
Policy & Patient & $w_{0\text{--}20}$ range & $t_{50}$ range & $t_{80}$ range & Default 95\% CI \\
\midrule
G2P2C & adolescent\#001 & .467--.511 & 15--20 & 40--40 & .471--.512 \\
G2P2C & adolescent\#004 & .289--.659 &  5--30 & 35--50 & .516--.615 \\
G2P2C & adult\#001      & .407--.726 &  5--25 & 25--45 & .624--.701 \\
G2P2C & adult\#004      & .430--.748 &  5--25 & 25--45 & .671--.730 \\
\addlinespace
PPO & adolescent\#001 & .320--.432 & 25--30 & 45--50 & .374--.471 \\
PPO & adolescent\#004 & .616--.649 & 10--10 & 30--30 & .623--.658 \\
PPO & adult\#001      & .530--.545 & 15--15 & 35--35 & .505--.579 \\
PPO & adult\#004      & .548--.634 & 10--15 & 30--35 & .591--.638 \\
\bottomrule
\end{tabular}
\end{table*}

\FloatBarrier
\section{Detailed Validation Breakdowns}
\label{app:validation_breakdowns}

\subsection{Individual-feature replacement}
At every decision, the 24 scalar inputs are ranked by absolute KernelSHAP importance. The prespecified background-mean intervention replaces either the top, random, or bottom $m\in\{4,8\}$ inputs. Table~\ref{tab:e6_breakdown} reports patient-level aggregates. Each patient contributes multiple rows, so the ordering counts are descriptive rather than independent-subject tests.

\begin{table}[t]
\centering
\caption{Individual-feature replacement on Scenario~1. Changes are mean absolute changes in normalized policy output across 20 patients per policy--mask-size row. Counts report patients for which top-ranked replacement exceeds the comparator.}
\label{tab:e6_breakdown}
\small
\begin{tabular}{lrrrrr}
\toprule
Policy & $m$ & Top & Random & Bottom & Top $>$ random / bottom \\
\midrule
G2P2C & 4 & .015179 & .003966 & .000053 & 20/20 / 20/20 \\
G2P2C & 8 & .014306 & .006953 & .000128 & 20/20 / 20/20 \\
PPO   & 4 & .006269 & .001602 & .000052 & 20/20 / 20/20 \\
PPO   & 8 & .007856 & .003107 & .000109 & 20/20 / 20/20 \\
\midrule
Pooled & -- & .010903 & .003907 & .000086 & 80/80 / 80/80 \\
\bottomrule
\end{tabular}
\end{table}

\subsection{Temporal-window replacement}
For each temporal intervention, only the four administered-insulin inputs in one 20-min window are replaced by their background means; all CGM inputs and all other insulin lags remain fixed. Table~\ref{tab:e7_breakdown} distinguishes highest-window agreement from full three-window ranking. Across 120 configurations, the highest-ranked window agrees in 115 cases, the mean Spearman correlation is 0.913, and the median is 1.0. The three-item rank correlations are $-0.5$ once, $0.5$ in 18 cases, and $1.0$ in 101 cases.

\begin{table}[t]
\centering
\caption{KernelSHAP versus direct window-replacement ranking across 20 patients per policy--scenario row. Agreement counts only whether the highest-ranked window matches. Spearman $\rho$ compares the complete three-window ranking.}
\label{tab:e7_breakdown}
\small
\begin{tabular}{llrrr}
\toprule
Policy & Scenario & Top-window match & Mean $\rho$ & Median $\rho$ \\
\midrule
G2P2C & 1 & 19/20 & .825 & 1.000 \\
G2P2C & 2 & 19/20 & .850 & 1.000 \\
G2P2C & 3 & 18/20 & .850 & 1.000 \\
PPO   & 1 & 20/20 & .975 & 1.000 \\
PPO   & 2 & 19/20 & .975 & 1.000 \\
PPO   & 3 & 20/20 & 1.000 & 1.000 \\
\midrule
All & All & 115/120 & .913 & 1.000 \\
\bottomrule
\end{tabular}
\end{table}

\subsection{KernelSHAP versus TimeSHAP}
TimeSHAP provides a sequential Shapley-method replication, not independent ground truth. Table~\ref{tab:e9_agreement} shows generally strong highest-window agreement, with 111/120 matches overall. Agreement is higher for PPO (58/60) than G2P2C (53/60). Table~\ref{tab:e9_profiles} shows that the exact magnitude difference is also larger for G2P2C: its recent fraction differs by 10.05 percentage points between methods, compared with 0.51 points for PPO.

\begin{table}[t]
\centering
\caption{KernelSHAP--TimeSHAP temporal-ranking agreement across 20 patients per policy--scenario row. Top-window agreement and full three-window Spearman correlation are distinct quantities.}
\label{tab:e9_agreement}
\small
\begin{tabular}{llrrr}
\toprule
Policy & Scenario & Top-window match & Mean $\rho$ & Median $\rho$ \\
\midrule
G2P2C & 1 & 16/20 & .750 & .750 \\
G2P2C & 2 & 20/20 & .850 & 1.000 \\
G2P2C & 3 & 17/20 & .775 & 1.000 \\
PPO   & 1 & 19/20 & .950 & 1.000 \\
PPO   & 2 & 20/20 & .975 & 1.000 \\
PPO   & 3 & 19/20 & .975 & 1.000 \\
\midrule
All & All & 111/120 & .879 & 1.000 \\
\bottomrule
\end{tabular}
\end{table}

\begin{table}[t]
\centering
\caption{Mean temporal profiles across 60 patient--scenario configurations per policy and method. Recent, middle, and older values are percentages normalized within administered-insulin history.}
\label{tab:e9_profiles}
\small
\begin{tabular}{llrrr}
\toprule
Policy & Method & Recent & Middle & Older \\
\midrule
G2P2C & KernelSHAP & 54.15 & 24.73 & 21.12 \\
G2P2C & TimeSHAP   & 64.20 & 20.73 & 15.07 \\
PPO   & KernelSHAP & 55.72 & 26.85 & 17.44 \\
PPO   & TimeSHAP   & 56.22 & 26.78 & 17.00 \\
\bottomrule
\end{tabular}
\end{table}

\FloatBarrier

\end{document}